\documentclass[conference]{IEEEtran}
\usepackage{times}

\usepackage[numbers]{natbib}
\usepackage{multicol}
\usepackage[bookmarks=true]{hyperref}
\usepackage{xcolor}  
\usepackage{tabularx}
\usepackage{multirow}
\usepackage{booktabs}
\usepackage{stfloats}
\usepackage{caption}
\usepackage{authblk}
\makeatletter
\renewcommand\AB@affilsepx{, \protect\Affilfont}
\makeatother

\usepackage{xcolor} 
\usepackage{hyperref} 
\usepackage{ulem} 

\definecolor{custompurple}{rgb}{0.4, 0.0314, 0.4549}  

\hypersetup{  
    colorlinks=true,
    linkcolor=custompurple,
    urlcolor=custompurple,
    citecolor=custompurple
}   

\usepackage[textsize=tiny]{todonotes}

\newcommand{\fuhang}[1]

\usepackage{graphicx}

\begin{document}

\title{FP3: A 3D Foundation Policy for Robotic Manipulation}

\author[*1\thanks{These authors contributed equally to this work.}]{Rujia Yang}
\author[*2,4$\dag$ \thanks{These authors contributed equally to this work.}]{Geng Chen}
\author[$\ddag$1,2,3]{Chuan Wen}
\author[$\ddag$1,2,3]{Yang Gao}
\affil[1]{IIIS, Tsinghua University}
\affil[2]{Shanghai AI Laboratory}
\affil[3]{Shanghai Qi Zhi Institute}
\affil[4]{UC San Diego}

\twocolumn[{%
\renewcommand\twocolumn[1][]{#1}%
\maketitle
\vspace{-0.5cm}
\begin{center}
    \centering
    \captionsetup{type=figure}
    \includegraphics[width=\textwidth]{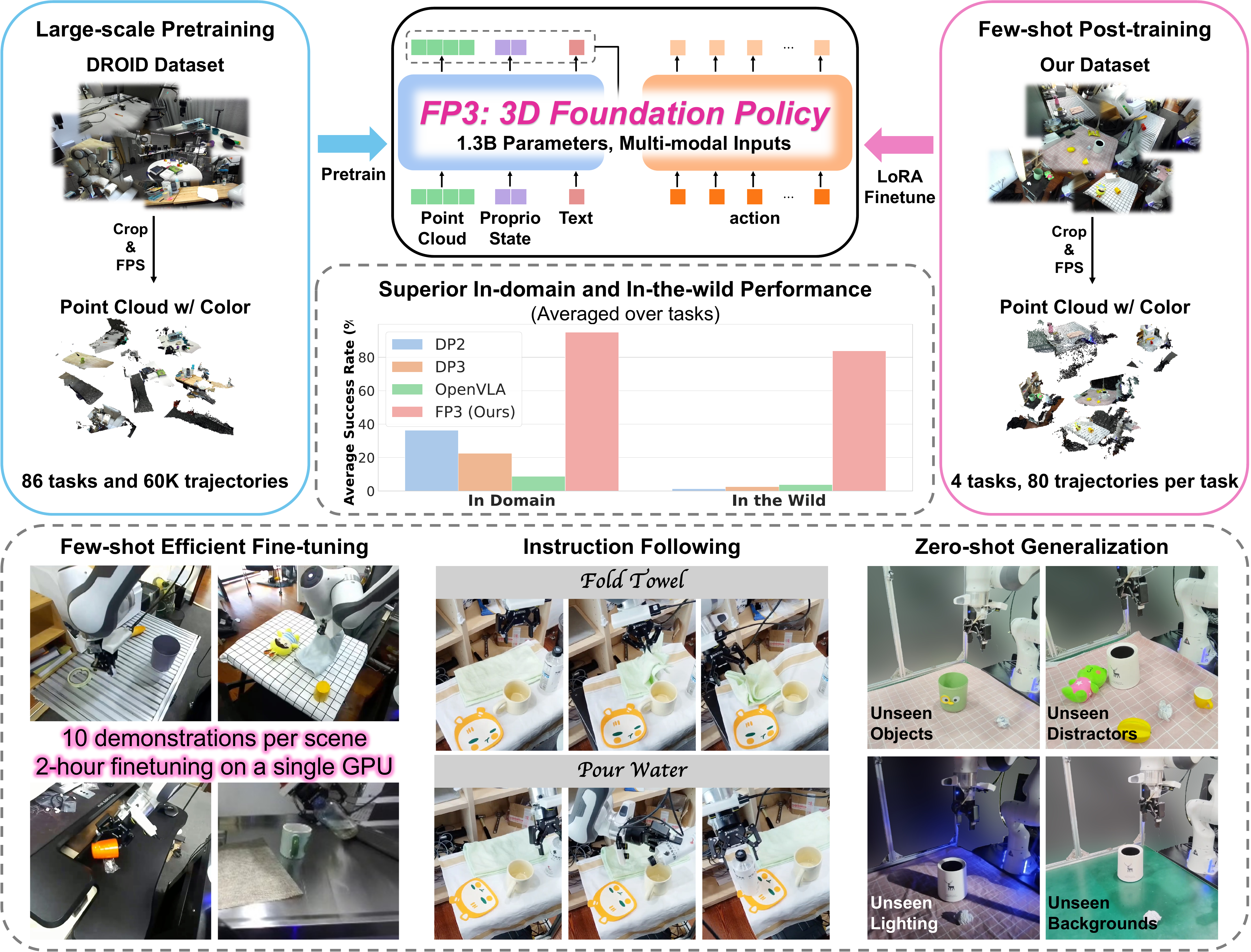}
        \captionof{figure}{\textbf{Overview of 3D Foundation Policy (FP3)}, a 1.3B 3D point cloud-based language-visuomotor policy pre-trained on 60k episodes from the DROID dataset \cite{droid}. FP3 supports data-efficient fine-tuning for downstream tasks, while demonstrating superior generalizability to unseen environments and novel objects.}
\end{center}%
}]
\vspace{0.5cm}
\makeatletter\def\Hy@Warning#1{}\makeatother\let\thefootnote\relax\footnotetext{$*$ denotes equal contribution. $\ddag$ denotes equal advising. $\dag$: work done during the internship at Shanghai AI Laboratory}

\begin{abstract}
Following its success in natural language processing and computer vision, foundation models that are pre-trained on large-scale multi-task datasets have also shown great potential in robotics. However, most existing robot foundation models rely solely on 2D image observations, ignoring 3D geometric information, which is essential for robots to perceive and reason about the 3D world. In this paper, we introduce FP3, a first large-scale 3D foundation policy model for robotic manipulation. FP3 builds on a scalable diffusion transformer architecture and is pre-trained on 60k trajectories with point cloud observations. With the model design and diverse pre-training data, FP3 can be efficiently fine-tuned for downstream tasks while exhibiting strong generalization capabilities. Experiments on real robots demonstrate that with only 80 demonstrations, FP3 is able to learn a new task with over 90\% success rates in novel environments with unseen objects, significantly surpassing existing robot foundation models. Visualizations and
code are available at: \href{https://3d-foundation-policy.github.io/}{FP3}.
\end{abstract}

\IEEEpeerreviewmaketitle

\section{Introduction}
Learning-based policies have shown great effectiveness in robotic manipulation \cite{rt1, rt2, dp, act, openvla, pi0}. However, these learned policies often show limited or even zero generalization capability to unseen scenarios, new objects, and distractors \cite{xie2024decomposing}. Additionally, most current methods are trained on single or few tasks\cite{dp,act}, requiring a relatively large amount of expert demonstrations (usually about 200 episodes) to learn a new task. In contrast, natural language processing (NLP) and computer vision (CV) have achieved remarkable success in developing foundation models that are trained on large-scale data and diverse tasks, enabling them to generalize to arbitrary scenarios in the wild. Therefore, building a similar foundation model in robotic manipulation that can generalize to novel objects, scenes, and tasks becomes a promising topic \cite{openvla, rdt, pi0}.

Towards this goal of policy foundation models, there have been some initial attempts at vision-language-action (VLA) models \cite{rt2, openvla, pi0}, which build upon the vision-language models (VLM) trained on internet-scale data of vision and language to inherit the commonsense knowledge and fine-tune the VLMs on large-scale robotics datasets \cite{oxe, droid}. In the meanwhile, works such as RDT \cite{rdt} try to scale up diffusion models to build foundation policies. Despite significant progress, their generalizability remains limited when confronted with novel tasks, objects, scenes, and camera views, etc.

One potential limitation of current policy foundation models is their exclusive reliance on 2D image observations, lacking 3D observation inputs. However, 3D geometric information is significant for perceiving 3D environments and reasoning about spatial relationships \cite{zhu2024point, sgr, dp3, wen2023can}. There have been works showing that 3D representations can improve the sample efficiency and generalizability of robotic manipulation policies \cite{peract, dp3, 3d_diffuser_actor, sgr}. Among all the 3D representations such as RGB-D images, point clouds, voxels, and 3D Gaussians \cite{3dgs}, point clouds are found to be the most effective \cite{dp3}.

In this work, we introduce 3D Foundation Policy (FP3), the first 3D point cloud-based language-visuomotor policy foundation model for robotic manipulation that exhibits strong generalizability and sample efficiency. To extract rich semantic and geometric representation from 3D point cloud observation, FP3 adopts a pre-trained large-scale point cloud encoder Uni3D \cite{uni3d}. We further leverage an encoder-decoder Diffusion Transformer (DiT) architecture to integrate the point cloud representations, language embedding, and proprioception for denoising the actions.

With the proposed policy architecture, we employ a pre-training\&post-training recipe for FP3, mirroring the common practice in large language models (LLMs) \cite{gpt4, gemini} where the model is pre-trained on large-scale diverse corpus and fine-tuned on curated task-specific data to adapt for downstream tasks. We first pre-train FP3 on the large-scale robotic manipulation dataset DROID \cite{droid} which contains 76k demonstration trajectories or 350h of interaction data from 564 scenes and 86 tasks. We then collect a small amount of high-quality teleoperation data for several tasks and fine-tune FP3. The result indicates that our model can efficiently master a new task with only 80 post-training trajectories and is capable of zero-shot generalization to both novel objects and environments with about 90\% success rate. In contrast, strong baselines like DP3 \cite{dp3} and OpenVLA \cite{openvla} almost completely fail in this setting. Due to the advantages of 3D representation,  FP3 is also robust to background variations, lighting conditions, camera angles, and distractors. Finally, we conduct ablation studies to demonstrate that the 3D representation, data scaling, and model scaling all contribute to the model's superior performance.

We summarize our main contributions as follows:

\begin{enumerate}
    \item We propose a novel diffusion-based 3D robot policy architecture, FP3.
    \item  We pre-train FP3 on large-scale robotic manipulation data with 3D observation, establishing a 1B-parameter 3D policy foundation model.
    \item We collect data on several new tasks and demonstrate the efficient and generalizable fine-tuning of FP3, achieving around 60\% in-domain and 80\% in-the-wild performance improvement on average over strong baselines with only 2-hour and single-GPU finetuning.
\end{enumerate}

\section{Related Work}

\subsection{Foundation Models in Robotics}
Similar to the cases in natural language processing and computer vision, foundation models have already been widely used in multiple aspects of robotics, including representation learning \cite{r3m, VIP,Wen2024atm,yuan2024generalflow}, high-level task planning \cite{saycan, vila}, training-free robotic manipulation \cite{voxposer, copa, rekep}, etc. In this work, we concentrate on policy foundation models in robotics, which also often refer to multi-task "generalist" robot policies \cite{rt1, rt2, openvla, octo, rdt, pi0} trained on large-scale robot datasets \cite{oxe, bridge, droid, rh20t}. A significant subset of policy foundation models consists of autoregressive vision-language-action (VLA) models like RT-2 \cite{rt2} and OpenVLA \cite{openvla}, which directly fine-tune pre-trained large vision-language models to predict actions by treating discretized actions as language tokens. Recently, RDT \cite{rdt} scales up a diffusion transformer \cite{dit} to build a foundation model for bimanual manipulation. Another recent work $\pi_0$ \cite{pi0} investigates the combination of a pre-trained VLM backbone with diffusion (flow matching \cite{flow_matching}) models for robot control. Our work is most similar to RDT in terms of architecture, as both build upon diffusion transformers, but there are key differences, such as conditioning blocks. RDT employs cross attention blocks, while we opt for adaLN blocks \cite{dit, dit_policy} for stabilizing training. 

In addition to architecture, a key difference between these approaches and FP3 is the observation modality. Unlike these works, which all take 2D image observation as input, our work utilizes 3D point cloud observations to enhance the perception of 3D geometric information and reasoning about spatial relationships, introducing the first policy foundation model with 3D representations to our knowledge.

\subsection{Robotic Manipulation with 3D representations}
Compared to 2D images, 3D representations such as RGB-D images, point clouds, and voxels contain richer geometric information and have thus been widely used in robotic manipulation \cite{peract, dp3, rvt}. Kite \cite{kite} directly leverages the RGB-D observation for semantic manipulation. Other works \cite{chen2023polarnet, yuanm2t2,dp3,rise} reconstruct point clouds from RGB-D images and process them using a point cloud encoder for manipulation. Voxelizing the point clouds for perception is also a viable solution \cite{james2022coarse, peract,huang2024fourier}. Another set of works\cite{gervet2023act3d,xian2023chaineddiffuser, 3d_diffuser_actor,sgr,sgrv2} lifts 2D image features to 3D space to benefit from both semantic and geometric information. There have also been attempts in combining implicit or explicit 3D reconstruction (NeRF \cite{nerf}, 3D gaussians \cite{{3dgs}}) with robotic manipulation \cite{ze2023gnfactor, dex, dai2023graspnerf, gaussiangrasper, lu2024manigaussian}. In this work, we choose the point cloud as the 3D representation as it is found to be more effective than other representations in DP3 \cite{dp3}.

Despite the differences in representation, all the aforementioned methods consist of small networks trained on a limited number of tasks. FP3 differs from these works in several ways: it is a foundation model trained on a large multi-task dataset, it scales up the point cloud encoder to 300M parameters and the whole network to 1.3B parameters, and it supports efficient and generalizable fine-tuning for new tasks.

\subsection{Diffusion models in Robotics}
Diffusion models have achieved great success in image generation \cite{ddpm, ddim, sd} and video generation \cite{vdm, svd, sora, gu2024seer} by modeling complex high-dimensional continuous distributions through progressive denoising. Due to their remarkable expressiveness, diffusion models have also been applied across various fields in robotics, including reinforcement learning \cite{wangdiffusion, ajayconditional, chenoffline}, imitation learning \cite{dp, dp3, mdt, rdt, octo, pi0}, and motion planning \cite{janner2022planning, urain2023se,carvalho2023mpd}. Our work focuses on ``Diffusion Policies" \cite{dp}, which refers to methods that directly employ conditional diffusion models as visuomotor policy models for imitation learning. Most closely related to our work is RDT \cite{rdt}, a diffusion foundation model for manipulation. A key distinction between our work and RDT is that our FP3 model leverages 3D point cloud representations to achieve improved data efficiency and generalizability.

\section{Method}
We introduce the 3D Foundation Policy (FP3) model for generalist robotic manipulation, achieving high data efficiency and generalization capability. FP3 is a 1.3B encoder-decoder transformer network following a two-stage pre-training and post-training recipe. We first provide the detailed architecture and key design decisions of FP3 in \autoref{subsec:model arch}. Then we describe the pre-training and post-training procedures in \autoref{subsec: pre-training} and \autoref{subsec:post-training}, respectively.

\begin{figure*}
    \centering
    \includegraphics[width=\linewidth]{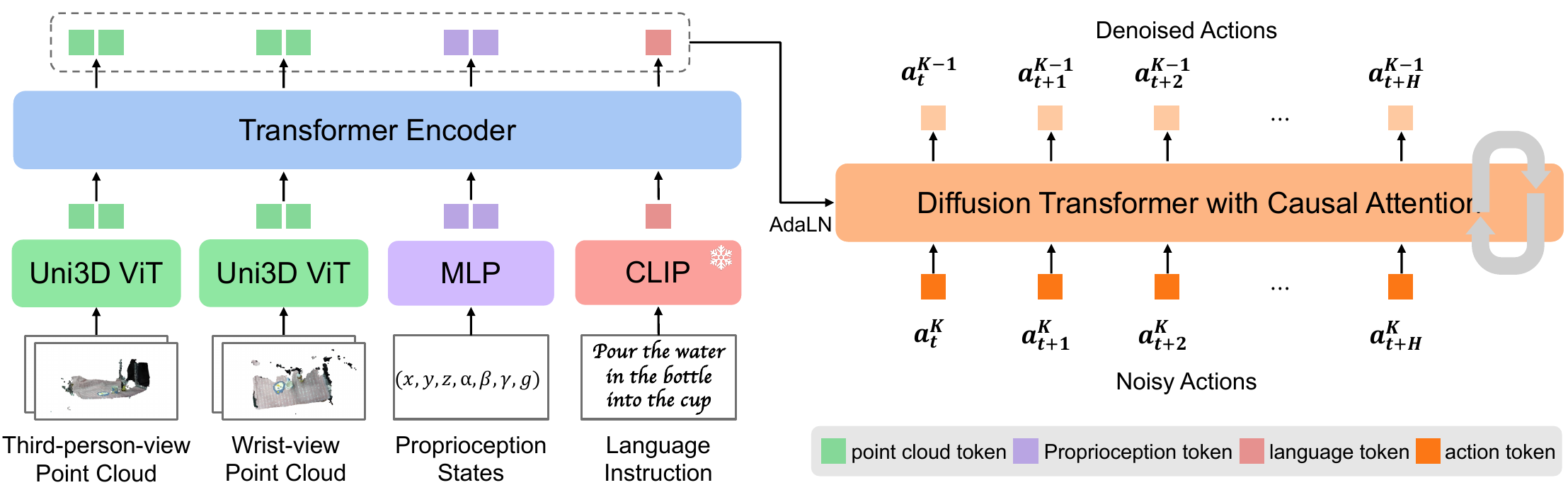}
    \caption{\textbf{FP3 architecture.} Each camera view's point cloud observation $\mathbf{P}_t^i$ 
 (with history length of two) is encoded with a Uni3D ViT-L \cite{uni3d} encoder. The language instruction $\ell_t$ is embedded with a frozen CLIP \cite{clip} model. The Transformer encoder fuses multi-modal input embeddings to latent tokens, while the Transformer decoder takes in the noise actions and leverages adaLN \cite{dit, brock2018large, karras2021style} blocks to integrate the latent tokens generated by the encoder, predicting denoised action chunks.}
    \label{fig:arch}
\end{figure*}

\subsection{FP3 model}
\label{subsec:model arch}

At its core, FP3 is a diffusion-based policy model similar to \cite{dp, rdt}. It takes the 3D point cloud observation, language, and robot proprioceptive state as input and predicts action chunks of future actions. Formally, we formalize the problem of language-conditioned visuomotor control as modeling the distribution $p(A_t|o_t)$, where $\mathbf{o}_t = [\mathbf{P}_t^1, ..., \mathbf{P}_t^n, \ell_t, \mathbf{q}_t]$ is the observation at time $t$ including point cloud observation $\mathbf{P}_t^i$ from the $i^{th}$ camera (including historical observation), language instruction $\ell_t$ and proprioceptive information $\mathbf{q}_t$ and $A_t = [a_t, a_{t+1}, ..., a_{t+H-1}]$ denotes the predicted action chunk. We train a denoising diffusion probabilistic model (DDPM) \cite{ddpm} to approximate the conditional distribution and use the denoising diffusion implicit model (DDIM) \cite{ddim} method to accelerate inference.

Now we describe the detailed structure of FP3 model, including the encoding of multi-modal inputs and the transformer-based encoder-decoder architecture.

\textbf{Encoding of multi-modal inputs.} To process the multi-modal input, we encode the input signals into a unified token space with the same dimensions as follows:

\begin{itemize}
    \item \textbf{Point cloud observations} contain rich semantic and geometric information and are found to be more suitable for policy learning compared to other 3D representations \cite{dp3}. Therefore, we consider using point cloud as the 3D representation in FP3. Current point cloud-based robot policies \cite{dp3, rise, sgrv2} typically use sparse point cloud and small networks such as PointNet++ \cite{pointnet++} and PointNeXt \cite{pointnext} to encode the points into embeddings. However, pre-trained large-scale foundation vision encoders have demonstrated a performance advantage over small encoders in image-based policies \cite{umi, data_scaling_law}. Consequently, we increase the number of input points to 4000 for each view and employ a 300M-parameter point cloud encoder Uni3D ViT \cite{uni3d} that is pre-trained to align the 3D point cloud features with the image-text aligned features to obtain the point cloud embeddings. For the third-person-view and the wrist-view point clouds, we use separate encoders since their point distributions might be greatly different. Following \cite{data_scaling_law}, we choose to fine-tune the weights of Uni3D ViTs during policy training.
    
    \item \textbf{Language instructions} are simply encoded with a CLIP \cite{clip} model to align with Uni3D. The weights are fixed during training since the language embeddings are already well-trained.
    
    \item \textbf{Low-dimensional inputs} including robot proprioceptive state and noise levels are processed with two-layer MLPs, respectively.
    
\end{itemize}

\textbf{Encoder-decoder structure.}
Since Diffusion Transformers have shown great scalability in image generation \cite{dit, sd3} and policy learning \cite{scale_dp, dit_policy},  we adopt the transformer architecture and scale it up for FP3. To better fuse the point cloud, language, and proprioceptive state embeddings, we utilize a Transformer Encoder-Decoder architecture similar to \cite{act, aloha_unleashed, mdt, dit_policy}. Specifically, FP3 first feeds all the embeddings into a transformer encoder, producing a sequence of informative latent tokens.

The diffusion denoiser of FP3 is a Transformer decoder that denoises the action chunks from noise with temporal causal masking following \cite{scale_dp}. To infuse the latent tokens with multi-modal information into the denoiser, FP3 adopts the adaptive Layer-Norm (adaLN) module for conditioning, which has been found to be essential for implementing diffusion training in image generation \cite{dit, sd3} and policy learning \cite{scale_dp, dit_policy}. 

\subsection{Pre-training}
\label{subsec: pre-training}

\textbf{Pre-training data.}
To build a 3D policy foundation model, we need to train our model on large-scale 3D robotic manipulation datasets. However, most existing large-scale robot datasets such as the Open X-Embodiment dataset (OXE, \cite{oxe}) are mainly 2D-only. Thus, in this work, we pre-train FP3 with the DROID dataset \cite{droid}, which includes 86 tasks and 76k demonstrations and provides depth observation data. We finally use 60k demonstrations from DROID to pre-train FP3.

\textbf{Data pre-processing.}
While DROID uses three cameras for data collection, we only use two of them including one third-view camera and one wrist-view camera in FP3 for convenience. We use the RGB image and the depth map to recover the 3D point cloud for each camera and transform the two point clouds to the same world frame. As we only care about the operated object, we cropped the points outside a 1-meter box to remove redundant points. Further, we downsample each point cloud by farthest point sampling (FPS, \cite{qi2017pointnet}) to 4000 points to facilitate model training while retaining sufficient information. We preserve the color channels of each point to enable further experiments conditioned on colors.

\textbf{Pre-training details.}
Following prior works \cite{openvla, data_scaling_law} which found that freezing pre-trained vision encoders may harm the policy performance, we fine-tune the Uni3D ViT encoder during pre-training. We also randomly drop some points during training for augmentation, and the dropout rate is randomly selected from 0 to 0.8.

We use the AdamW optimizer \cite{adamw} with a cosine learning rate schedule. The weight decay is set to 0.1, and gradient clipping is set to 1.0. The FP3 base model is pre-trained for 3M steps with a batch size of 128 using 8 NVIDIA A800 GPUs, which takes about 48 hours. Fine-tuning the same model on a single NVIDIA A800 GPU takes approximately 2 hours and can be further sped up with multi-GPU training.

To handle the partial observation, we stack 2 frames as input, including 1 step observation history, to compensate for the missing dynamic information of the robot.

\subsection{Post-training}
\label{subsec:post-training}

After obtaining the pre-trained base model, we further employ a post-training process using a small amount of high-quality data to adapt the model to certain tasks, which aligns with most modern LLM practice \cite{gpt4, gemini}. Different from the fine-tuning settings adopted in most existing robot foundation models in which they focus on either fine-tuning the model to adapt to new robot setups \cite{openvla, octo} or learning new tasks in a fixed environment \cite{rdt, pi0}, our goal is to fine-tune our model to \textit{solve specific tasks on any object, in any environment}.

To achieve this goal, we further collect data for each downstream task in our robot setup. Taking the lesson from \citet{data_scaling_law}, we aim to enhance the diversity of environments and objects rather than merely increasing the number of demonstrations in the same scenario. Specifically, for each task, we collect 10 teleoperation demonstrations in each of 8 environments using 8 unique objects, i.e., 80 demonstrations in total. We then fine-tune the base model on this data using the parameter-efficient fine-tuning strategy LoRA \cite{lora}. Thanks to the effective initialization from pre-training, this small amount of fine-tuning data enables zero-shot deployment to novel environments and objects. We will discuss the tasks and results in detail in \autoref{sec: experiments}.

\section{Experiments}
\label{sec: experiments}
We conduct experiments on real robots for four downstream tasks to investigate the following questions:

\begin{enumerate}
    \item Can FP3 be efficiently fine-tuned for new tasks?
    \item How well does the fine-tuned FP3 generalize to unseen objects and scenes compared to the existing imitation learning (foundation) policies?
    \item How robust is FP3 to the environment perturbations, such as lighting, camera views, distractors, \textit{etc.}? 
    \item  Can FP3 correctly execute the corresponding tasks following the language instruction?
\end{enumerate}

\subsection{Experimental Setups}
\noindent\textbf{Real robot setup.}
As we pre-train our FP3 model on the DROID dataset, we also build a real robot setup similar to DROID for evaluating downstream tasks. This setup features a Franka Emika Panda robot arm equipped with a Robotiq gripper, mounted on a movable desk. For point cloud observation, we utilize one ZED mini camera in wrist view and one ZED 2 camera in third-person view. To collect data, we employ a Meta Quest 2 VR headset to teleoperate the robot. We record the absolute Cartesian Space Control as the action space for policy training and deployment. More details are provided in the Appendix.

\begin{figure}[ht]
    \centering
    \includegraphics[width=\linewidth]{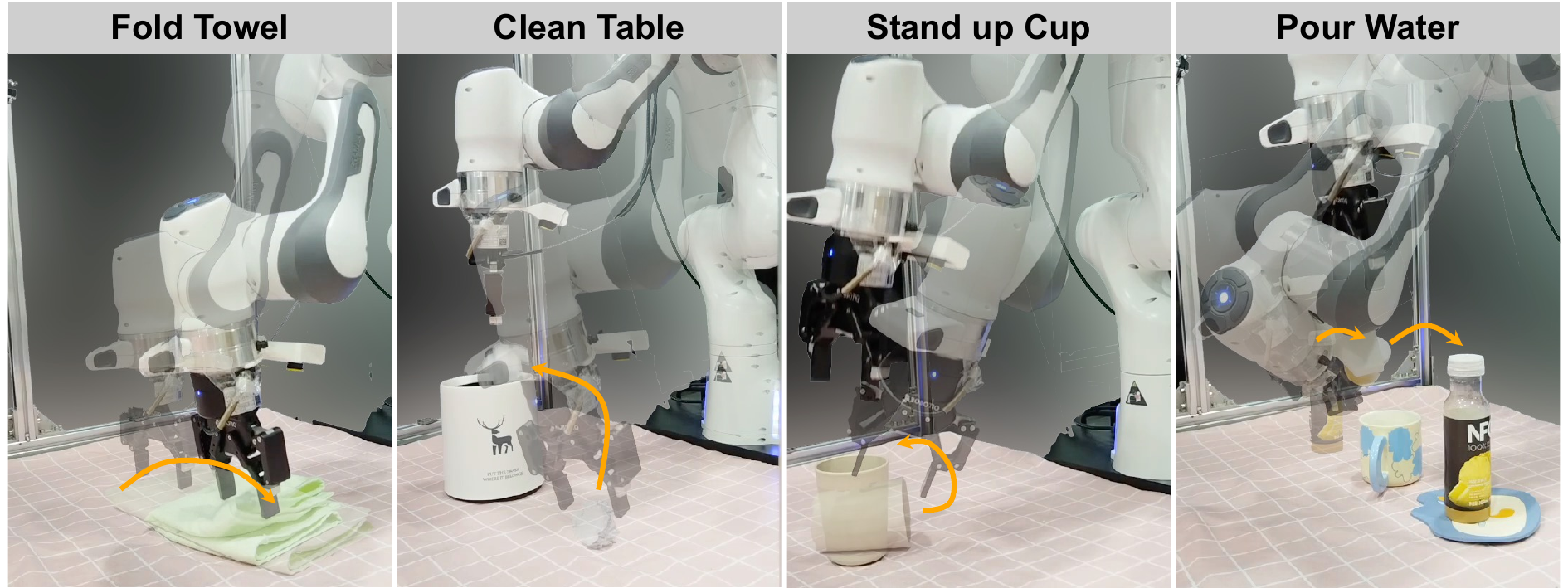}
    \caption{\textbf{Task illustrations.} We evaluate our model on four downstream tasks: \texttt{Fold Towel}, \texttt{Clean Table}, \texttt{Stand up Cup}, and \texttt{Pour Water}.}
    \label{fig:task}
\end{figure}

\noindent\textbf{Tasks.} We choose four downstream tasks to evaluate our model and the baselines:
\begin{itemize}
    \item \texttt{Fold Towel}: Fold a long towel from right to left on a flat surface.
    \item  \texttt{Clean Table}: Pick up a crumpled piece of paper and put it in the bucket.
    \item \texttt{Stand up Cup}: Stand a cup lying on a flat surface upright.
    \item \texttt{Pour Water}: Pick up a water bottle, pour the water from the water bottle into a cup, then place the water bottle on a coaster.
\end{itemize}

\autoref{fig:task} illustrates the process of these four tasks. Further details are provided in the Appendix.

\noindent\textbf{Baselines.}
In order to comprehensively evaluate FP3, we carefully select three baselines:
\begin{itemize}
    \item Diffusion Policy (DP) \cite{dp}: a classic diffusion-based imitation learning policy with 2D image observation.
    \item DP3 \cite{dp3}: an alternative version of DP which changes the 2D image observation to 3D point cloud and designs a lightweight encoder to encode the point cloud.
    \item OpenVLA \cite{openvla}: a most widely-used image-based Vision-Language-Action (VLA) model.
\end{itemize}

These three baselines each represent a small 2D policy, a small 3D policy, and a large 2D foundational policy. For DP and DP3, we add a language-conditioning module in the same manner as FP3 to fuse language instructions.

\noindent\textbf{Metrics.}
We report the success rate as our metric. Results in \autoref{tab:fine-tuning evaluation} are averaged over 20 evaluation trials.

\subsection{Efficient and generalizable fine-tuning for new tasks}
We first evaluate FP3's capability to learn new tasks efficiently. We collect only 10 demonstrations per environment-object pair for 8 pairs to obtain 80 demonstrations in total for each task. For DP and DP3, we use the demonstrations to train the policies, while for OpenVLA and our FP3, we follow the pre-training and post-training recipe to fine-tune the base models for each task. Additionally, we train FP3 from scratch to validate the necessity of pre-training.

For each task, we not only evaluate all the policies in four in-domain environments with seen objects, but also deploy them zero-shot in four out-of-domain environments with unseen objects, which is a huge challenge for the model's generalizability.

\textbf{In-domain Performance.} Results in \autoref{tab:fine-tuning evaluation} show that in in-domain experiments, with only 10 demonstrations per scene, DP and DP3 can somewhat handle easier tasks, even though the success rate is below 50\% in most cases; however, they almost completely fail in the more difficult task \texttt{Pour Water}. And OpenVLA struggles to perform any task, likely due to the lack of action chunking. In contrast, thanks to pre-training and 3D representation, FP3 efficiently learns all tasks with a success rate exceeding 90\%. Qualitatively, we find that the failures of all baseline algorithms are mainly due to issues in the details, such as not being precise enough when attempting to grasp objects, which causes the object to be pushed away, or the bottle opening being off-center when pouring water, etc. In contrast, due to its large number of parameters and extensive pre-training, our FP3 policy can better predict complex target actions. The actions predicted by the FP3 policy are significantly smoother and more precise, leading to a notably higher success rate compared to the strong baselines.

\begin{figure*}
    \centering
    \includegraphics[width=\linewidth]{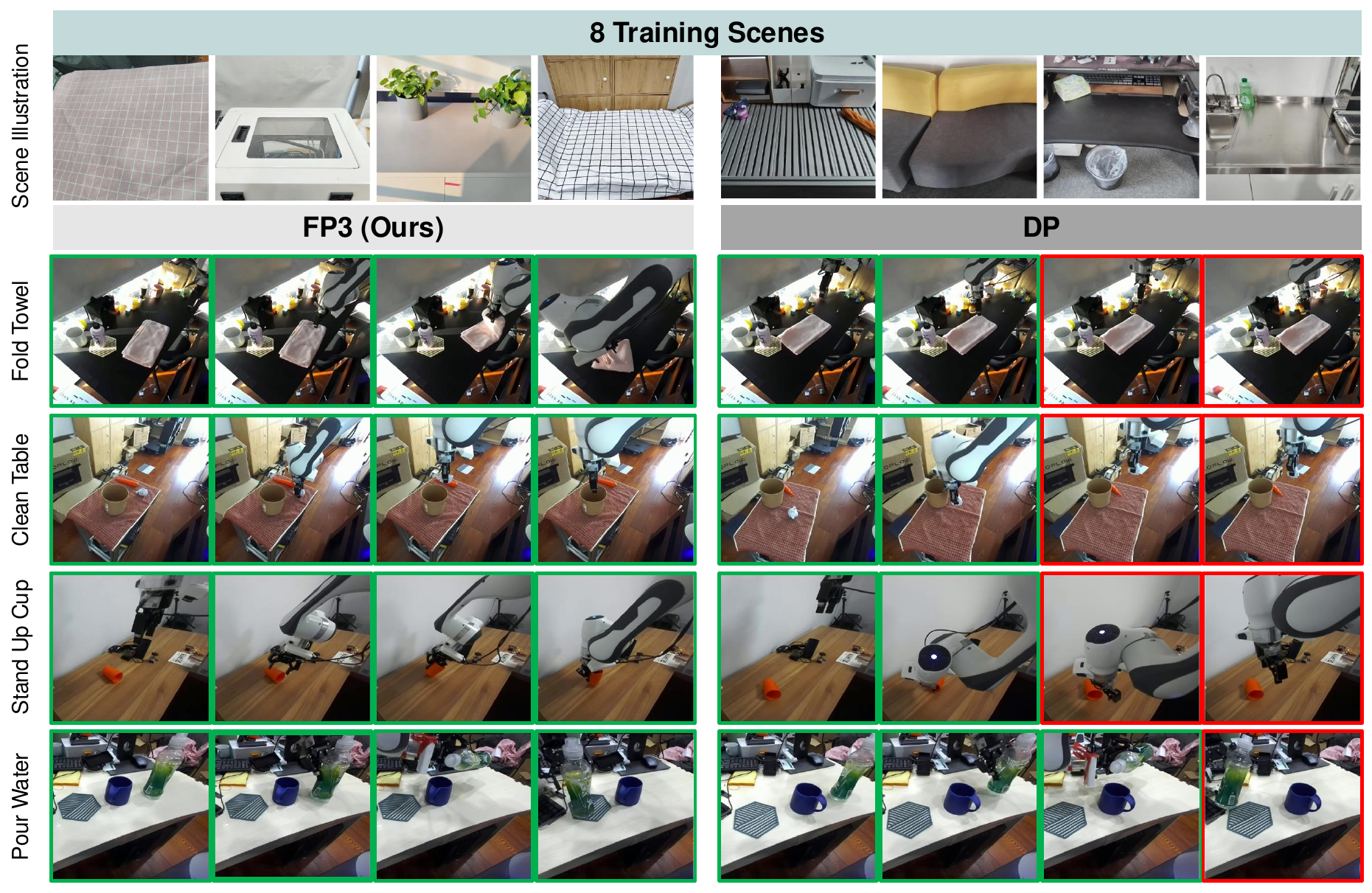}
    \caption{\textbf{Visualizations of post-training environments and in-the-wild evaluations.} The green boxes represent successful steps, while the red boxes represent failed ones. FP3 generalize well to all unseen environments and new objects, while Diffusion Policy  often fails to recognize the target object or misses the target position.}
    \label{fig:vis}
\end{figure*}

\textbf{In-the-Wild Performance.} We further move the robot arm to novel environments and evaluate the policies with unseen objects. In this challenging setting, we observe that all baseline policies without pre-training, including FP3-Scratch, often fail to recognize the target objects, resulting in near-zero performance as shown in \autoref{fig:vis}. In contrast, FP3 rarely encounters such situations and consistently performs well in all scenarios and tasks, achieving an average success rate of over 80\%, which devastates all the baselines. We attribute the superior performance to our large-scale pre-training, as the pre-training data encompasses a wide variety of scenes and objects, greatly enhancing the robustness of the policy. Furthermore, the point cloud observation is also a crucial factor, enabling better capture of geometric information, which is essential for cross-domain generalization.

\textbf{Failure Analysis of Baselines.} It is worth noting that in all cases, OpenVLA performs poorly, with the primary failure modes being its tendency to get stuck at specific positions and its inability to interact accurately with objects. The issue of getting stuck may stem from OpenVLA's lack of action chunking and observation history. The failure to interact precisely might be because OpenVLA uses only third-person-view observation, which offers a restricted field of view.

Another interesting issue is the policy’s response after an initial failure attempt. Sometimes, the policy fails in its first attempt, as seen in the \texttt{Clean Table} task where it fails to grasp the crumpled paper and ends up grabbing nothing. In such cases, we observe that only OpenVLA and FP3 make reasonable subsequent attempts, while DP, DP3, and FP3-Scratch tend to get stuck and simply wobble around the area after failure, unable to have a try again. This phenomenon happens probably because the fine-tuning data is limited, thus the policies without pre-training can fall into an out-of-distribution state after the first failure, and hence fail to output reasonable behaviors. Conversely, Large-scale pre-training with diverse tasks and objects in FP3 addresses this issue.

\begin{table*}[htbp]
\centering
\caption{\textbf{Post-training Evaluation.} We fine-tune FP3 and baseline methods on 80 demonstrations from 8 environments and evaluate them on four in-domain environments with seen objects and four in-the-wild environments with unseen objects, conducting 5 trials for each. FP3 significantly outperforms other policies both in domain and in the wild.}  
\small  
\setlength{\tabcolsep}{3pt} 
\newcolumntype{C}{>{\centering\arraybackslash}X} 
\begin{tabularx}{\textwidth}{lCC CC CC CC CC} 
\toprule[1.5pt]  
\textbf{} & \multicolumn{2}{c}{Fold Towel} & \multicolumn{2}{c}{Clean Table} & \multicolumn{2}{c}{Stand up Cup} & \multicolumn{2}{c}{Pour Water} & \multicolumn{2}{c}{Average} \\   
\cmidrule(lr){2-3} \cmidrule(lr){4-5} \cmidrule(lr){6-7} \cmidrule(lr){8-9} \cmidrule(lr){10-11} 
                 & In-domain & In-the-wild & In-domain & In-the-wild & In-domain & In-the-wild & In-domain & In-the-wild & In-domain & In-the-wild \\

\midrule  
DP        &    20              &      0             &      75             &     5            &         45          &      0            & 5 & 0 & 36.25 & 1.25 \\ 

DP3        &     20             &       0            &        25         &       10           &          45         &       0            & 0 & 0 & 22.50 & 2.50 \\ 
OpenVLA       &     5             &       0          &      15         &       5           &    15             &          10         & 0 & 0 & 7.50 & 3.75 \\ 
\midrule  
FP3-Scratch       &          35         &       5            &       35           &   0                &    15              &     0             & 35 & 0 & 30.00 & 1.25 \\ 
FP3 (\textbf{ours})       &      \textbf{90}            &       \textbf{85}           &       \textbf{100}            &    \textbf{95}              &       \textbf{95}          &      \textbf{75}              & \textbf{95} & \textbf{75} & \textbf{95.00} & \textbf{82.50} \\ 
\bottomrule[1.5pt]  
\end{tabularx}   
\label{tab:fine-tuning evaluation}  
\end{table*}

\subsection{More experiments on generalization}

\begin{figure*}
    \centering
    \includegraphics[width=\linewidth]{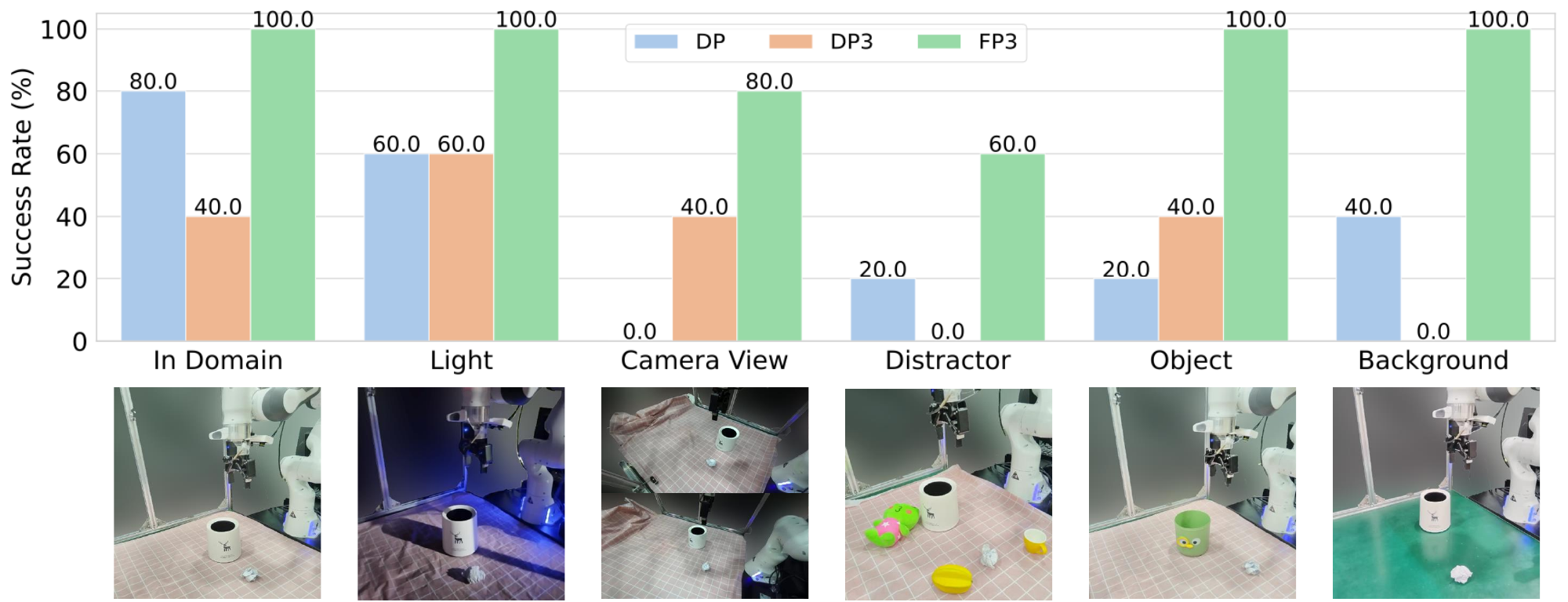}
    \caption{\textbf{Generalization evaluation.} We evaluate FP3 and baseline policies on a diverse set of tasks, covering different axes of generalization, including lighting, camera view, distractor, object and background. FP3 achieves outstanding performance in all generalization evaluation settings.}
    \label{fig:generalization}
\end{figure*}

Having demonstrated the efficient adaptation to new tasks and remarkable generalizability to novel objects and environments of FP3, we conduct more comprehensive experiments on FP3's generalizability to different environments and robot setups using the \texttt{Clean Table} task. \autoref{fig:generalization} demonstrates the results and visualizations.

\textbf{Generalization to different object appearances, backgrounds, and lighting conditions.}
Conventional image-based policy networks are sensitive to visual variations. Therefore, we systematically alter one aspect of the in-domain environment: object appearance, background texture, or lighting condition, to evaluate the policies. The results indicate that the image-based DP experiences a performance decline compared to the in-domain results, particularly with modified color and background. In contrast, DP3's performance in lighting and object color generalization remains stable, as it eliminates the color channels and relies solely on point positions. However, it is still limited by in-domain performance. With the benefits of pre-training initialization and 3D geometry understanding, FP3 surpasses the baselines. Qualitatively, DP and DP3 occasionally struggle to accurately recognize objects or estimate their positions, while FP3 consistently performs well.

\textbf{Generalization to new camera views.}
Camera view variations have also been a major challenge for image-based policies. Therefore, we adjust the camera view by approximately 30 degrees from the training data to evaluate the robustness of the policies. Once again, DP completely fails in this scenario, and DP3 is constrained by its in-domain performance, while FP3 maintains its high performance since the point cloud is consistently converted to the same coordinates as long as the camera is properly calibrated.

\textbf{Generalization to distractors.}
We also try putting random distractors around the target object to assess the robustness of these policies. In such settings, we find that the policies may attempt to grasp interfering objects. This issue occurs in all methods, including FP3, yet FP3's performance remains the highest and most stable.

\subsection{Instruction following}

\begin{figure*}
    \centering
    \includegraphics[width=\linewidth]{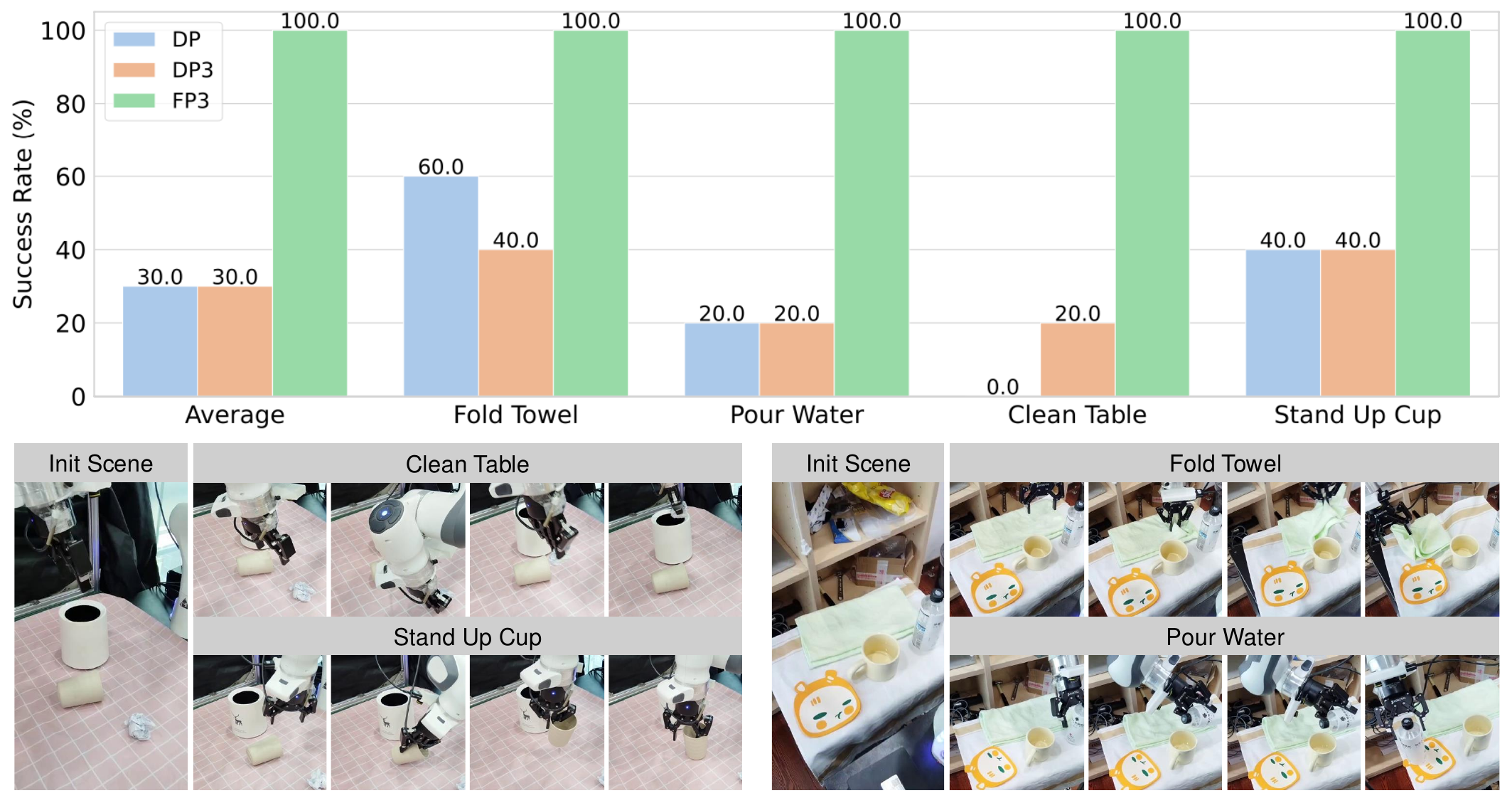}
    \caption{\textbf{Instruction following evaluation.} We evaluate FP3 and baseline policies in the same initial state with different language instructions. FP3 can perfectly follow the instructions to execute the correct tasks rather than simply memorize the training distribution.}
    \label{fig:language}
\end{figure*}

Since FP3 is a language-conditioned visuomotor policy, it's also important to evaluate its capability to execute tasks following the language command. Hence we fine-tune FP3 and the baselines in a multi-task setting using data from all tasks and evaluate them by providing different language instructions in the same initial state. \autoref{fig:language} demonstrates that FP3 can execute tasks according to different language commands within the same starting context, while the baseline methods either fail to complete the task or are disrupted by the target objects of other tasks.

\subsection{Ablations}
We finally do ablation studies on the observation choice, model size, and pre-training data size. We consider the following variants of FP3:
\begin{itemize}
    \item FP3-Base reduces the transformer size of both the encoder and decoder in FP3 from ViT-Large to ViT-Base, decreasing the total parameters from 1.3B to 365M.
    \item FP3-Base-30k further reduces the pre-training data of FP3-Base from 60k to 30k demonstrations.
    \item FP3-Base-Image converts the point cloud observations to image observations on FP3-Base and employs a DINOv2 \cite{dinov2} model to encode the images.
    \item FP3-Scratch is the FP3 model trained from scratch without any pre-training.
\end{itemize}

\autoref{tab:ablations} presents the results of each variant on the \texttt{Clean Table} task. FP3-Scratch exhibits poor performance both in domain and in the wild, indicating the importance of pre-training. FP3-Base-Image's performance aligns with FP3-Base in domain, but there is a huge performance drop in the wild, highlighting the effectiveness of 3D point cloud representation. FP3-Base and FP3-Base-30k demonstrate similar performance, both lower than our final FP3. More challenging tasks and additional data points are likely required to draw more definitive conclusions about the precise scaling law of pre-training.

\begin{table}[htbp]  
\caption{\textbf{Ablation study.} We achieve the best performance when using 3D point cloud input, a larger model, and larger-scale pre-training data.}   
\small  
\setlength{\tabcolsep}{3pt} 
\newcolumntype{C}{>{\centering\arraybackslash}X} 
\begin{tabularx}{\columnwidth}{lCC} 
\toprule[1.5pt]  
\textbf{} & \multicolumn{2}{c}{Clean Table} \\   
\cmidrule(lr){2-3} 
                 & In-domain & In-the-wild \\

\midrule  
FP3-Scratch        &      35             &     0            \\  

FP3-Base-Image       &       90           &   55 
           \\  
FP3-Base      &      95         &       90           \\  
FP3-Base-30k        &        95         &       90
               \\  
\midrule
FP3 (\textbf{ours})       &       \textbf{100}            &    \textbf{95}              \\  
\bottomrule[1.5pt]  
\end{tabularx}   
\label{tab:ablations}  
\end{table}
\section{Limitations}
While FP3 shows strong performance as a policy foundation model, it still has several limitations. One limitation is that although FP3 enables efficient and generalizable downstream fine-tuning, the base model exhibits limited zero-shot performance. One possible reason is that the pre-training dataset DROID is still not large enough compared to other 2D robotics datasets like OXE. Future work can consider collecting larger 3D robotics datasets for pre-training. Another limitation is that FP3 incorporates language conditioning through a simple CLIP embedding, which is insufficient to represent complicated and dynamic information. Combining diffusion-based FP3 with VLM to build a VLA model like $\pi_0$ \cite{pi0} seems a promising future direction. Additionally, FP3 does not leverage the robust pre-trained 2D vision encoders like DINOV2 \cite{dinov2} and SigLIP \cite{siglip}. There is huge potential in merging 3D point cloud features with 2D image features or lifting the 2D features to 3D space. We leave these explorations for future work.

\section{Conclusion}
In this work, we present the 3D Foundation Policy (FP3), a large-scale Diffusion Transformer-based policy with 3D point cloud input. We pre-train FP3 on 60k episodes of robotic manipulation data and subsequently fine-tune it for downstream tasks. Through extensive experiments, we demonstrate that FP3 serves as an outstanding policy initialization for data-efficient and generalizable fine-tuning for new tasks. With only 80 demonstrations, FP3 can learn a new task with over 90\% success rates in novel environments with unseen objects, significantly outperforming existing robot policies. We hope that our work will pave the way for more exciting advancements in robot foundation models utilizing 3D representations.

\section{Acknowledgment}
This work is supported by the National Natural Science Foundation of China (62176135), National Key R\&D Program of China (2022ZD0161700), Shanghai Qi Zhi Institute Innovation Program SQZ202306 and the Tsinghua University Dushi Program.

We would like to express our gratitude to Tong Zhang and Yingdong Hu for their help with the robot hardware setup.. We also appreciate the insightful discussions and feedback from Jiacheng You, Shengjie Wang and Chengbo Yuan.





\bibliographystyle{plainnat}
\bibliography{references}

\clearpage
\appendix

\section{Environment and Object Visualizations}
\subsection{Environment Visualization}
\begin{figure*}[h]
    \centering
    \includegraphics[width=\linewidth]{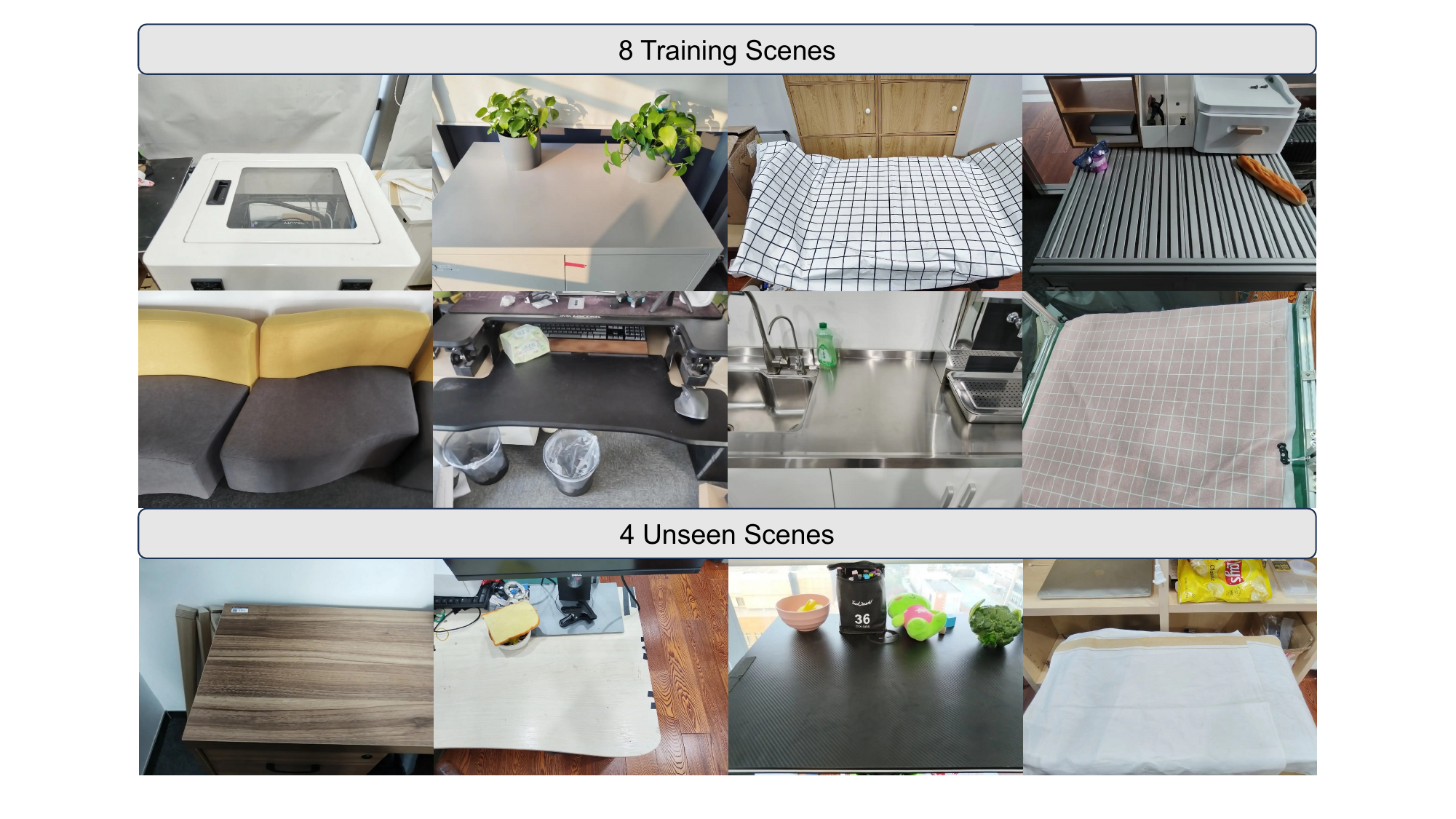}
    \caption{\textbf{Scenes visualization} of 8 post-training scenes and 4 unseen scenes for evaluation.}
    \label{fig:scenes}
\end{figure*}

For all four tasks, we collect post-training data in 8 environments and evaluate the policies in 4 in-domain environments and 4 unseen environments. We visualize all the scenes in Figure \ref{fig:scenes}.

\subsection{Object Visualization}
\begin{figure*}
    \centering
    \includegraphics[width=\linewidth]{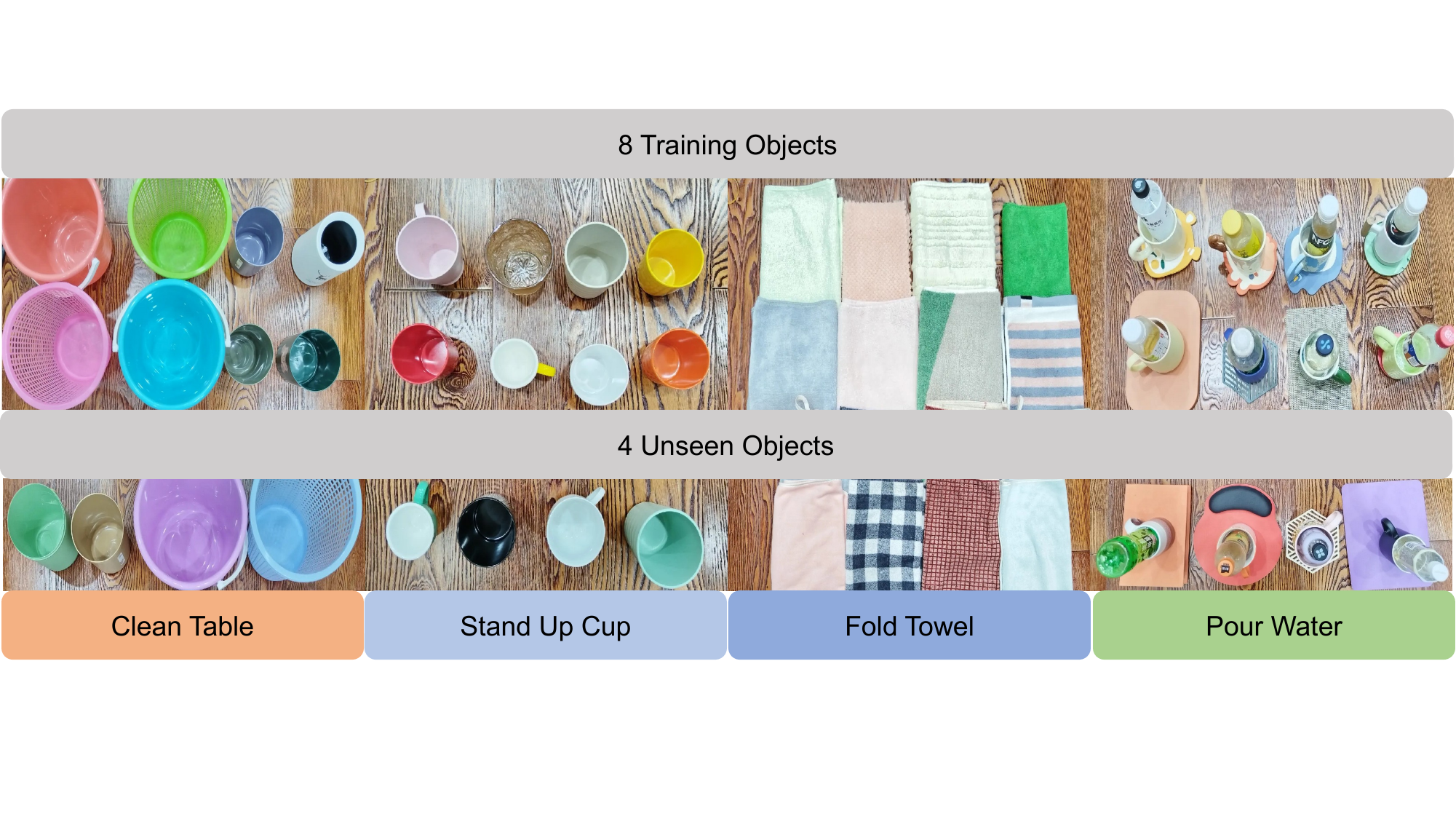}
    \caption{\textbf{Object Visualization} of 8 training objects and 4 unseen objects for each task.}
    \label{fig:objects}
\end{figure*}

Similar to the scenes, we have 8 training objects and 4 unseen objects for each task. All the objects are visualized in Figure \ref{fig:objects}.

\section{Training details}
We list the training hyperparameters for pre-training and fine-tuning in \autoref{tab:pretraining_hyper} and \autoref{tab:finetuning_hyper}. We train
all models on 8 NVIDIA A800 GPUs.

\begin{table}[htbp]  
\caption{\textbf{Pre-training Hyperparameters.}}   
\small  
\setlength{\tabcolsep}{3pt} 
\newcolumntype{C}{>{\centering\arraybackslash}X} 
\begin{tabularx}{\columnwidth}{lCC} 
\toprule[1.5pt]  
{Hyperparameter} & \multicolumn{1}{c}{Value} \\   

\midrule  
Image observation horizon        &      2            \\  

Proprioception observation horizon       &       2         \\  
Action prediction horizon      &      16        \\  
Action execution horizon        &        8        \\
Pointcloud down sample number & 4000 \\
Optimizer      &   AdamW \\
Learning rate & 1e-4 \\
Learning rate schedule & cosine \\
Weight decay & 0.1 \\
Gradient clipping & 1.0 \\
Batch size & 128 \\
Inference denoising iterations & 16 \\
\bottomrule[1.5pt]  
\end{tabularx}   
\label{tab:pretraining_hyper}  
\end{table}

\begin{table}[htbp]  
\caption{\textbf{Fine-tuning Hyperparameters.}}   
\small  
\setlength{\tabcolsep}{3pt} 
\newcolumntype{C}{>{\centering\arraybackslash}X} 
\begin{tabularx}{\columnwidth}{lCC} 
\toprule[1.5pt]  
{Hyperparameter} & \multicolumn{1}{c}{Value} \\   

\midrule  
LoRA rank & 32 \\
LoRA alpha & 16 \\
Image observation horizon        &      2            \\  

Proprioception observation horizon       &       2         \\  
Action prediction horizon      &      16        \\  
Action execution horizon        &        8        \\
Pointcloud down sample number & 4000 \\
Optimizer      &   AdamW \\
Learning rate & 1e-6 \\
Learning rate schedule & cosine \\
Weight decay & 0.1 \\
Gradient clipping & 1.0 \\
Batch size & 128 \\
Inference denoising iterations & 16 \\
\bottomrule[1.5pt]  
\end{tabularx}   
\label{tab:finetuning_hyper}  
\end{table}

\section{Task details}
\begin{itemize}
    \item \texttt{Fold Towel}: The robot is required to complete two steps: first, grasping the right edge of the towel, and second, folding the towel to the left. The towel's center position is randomly varied, and its orientation is randomly rotated within ±30 degrees. The towels have different colors, textures, and materials, but they are pre-folded into approximately rectangular shapes of similar size.
    \item \texttt{Clean Table}: The robot is required to complete two steps: first, grasping the spitball on the surface, and second, moving it just above the trash bin and dropping it into the bin. The positions of the spitball and the trash bin are randomly placed. The trash bins have different colors and textures, and their sizes vary to some extent.
    \item \texttt{Stand up Cup}: The robot is required to complete two steps: first, it inserts the gripper into the cup's opening and grasps the cup, and second, it lifts and stands the cup upright. The cup's placement is random, with its orientation varying within the 180-degree range where the opening faces the robotic arm.
    \item \texttt{Pour Water}: The robot performs three sequential actions: initially, it grasps a drink bottle;
    subsequently, it pours water into a mug; and finally, it places the bottle on a coaster. The position of the mug is randomly varied. The relative positions are also randomized, but the water bottle is always approximately to the left of the cup, while the coaster is approximately to the right of the cup. The water bottle, cup, and coaster vary in color, material, and size.
\end{itemize}
\section{Hardware setup}
As Figure \ref{fig:hardware} shows, we basically setup the robot following the DROID. We use a Franka
Emika Panda robot arm equipped with a Robotiq gripper,
mounted on a movable lifting desk. For point cloud observation, we
use one ZED mini camera and one ZED 2 camera. The ZED 2 provides third-person perspective images, fixed on a movable lifting desk, while the ZED mini provides wrist perspective images, fixed on the camera's end effector. For data
collection, we use a Meta Quest2 VR glass to teleoperate the
robot with a control frequency of 15Hz. All policy evaluations are performed on an RTX 3090 GPU (24GB VRAM). And everything is powered by a mobile power supply (EcoFlow DELTA 2 Max). 
\begin{center}
    \centering
    \captionsetup{type=figure}
    \includegraphics[width=0.5\textwidth]{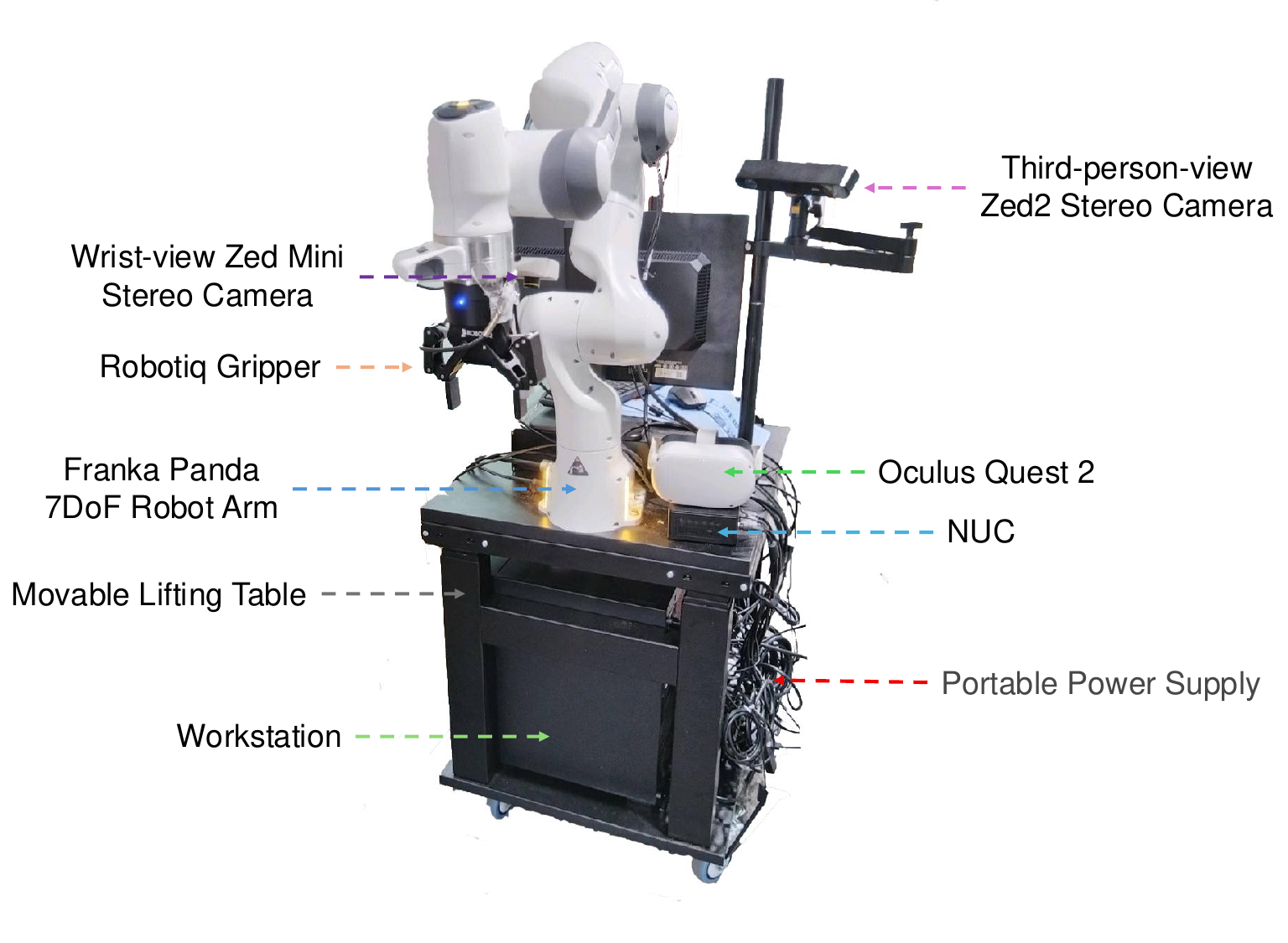}
    \captionof{figure}{hardware setup.}
    \label{fig:hardware}  
\end{center}%

\end{document}